\documentclass{article}

\usepackage[preprint]{neurips}

\usepackage[utf8]{inputenc} 
\usepackage[T1]{fontenc}    
\usepackage{hyperref}       
\usepackage{url}            
\usepackage{booktabs}       
\usepackage{amsfonts}       
\usepackage{nicefrac}       
\usepackage{microtype}      
\usepackage{xcolor}         
\usepackage{enumitem} 
\usepackage{multirow}
\usepackage{graphicx}
\usepackage{adjustbox}
\usepackage{cleveref} 
\usepackage[normalem]{ulem}

\title{Text-based Talking Video Editing with Cascaded Conditional Diffusion}

\author{%
  Bo Han \\
  Zhejiang University \\
  \And
  Heqing Zou \\
  Nanyang Technological University \\
  \And
  Haoyang Li \\
  Nanyang Technological University \\
  \And
  Guangcong Wang \\
  Nanyang Technological University \\
  \And
  Chng Eng Siong \\
  Nanyang Technological University \\
}

\begin{document}

\maketitle

\begin{abstract}
Text-based talking-head video editing aims to efficiently insert, delete, and substitute segments of talking videos through a user-friendly text editing approach. It is challenging because of \textbf{1)} generalizable talking-face representation, \textbf{2)} seamless audio-visual transitions, and \textbf{3)} identity-preserved talking faces. Previous works either require minutes of talking-face video training data and expensive test-time optimization for customized talking video editing or directly generate a video sequence without considering in-context information, leading to a poor generalizable representation, or incoherent transitions, or even inconsistent identity.
%
%
In this paper, we propose an efficient cascaded conditional diffusion-based framework, which consists of two stages: audio to dense-landmark motion and motion to video. \textit{\textbf{In the first stage}}, we first propose a dynamic weighted in-context diffusion module to synthesize dense-landmark motions given an edited audio. \textit{\textbf{In the second stage}}, we introduce a warping-guided conditional diffusion module. The module first interpolates between the start and end frames of the editing interval to generate smooth intermediate frames. Then, with the help of the audio-to-dense motion images, these intermediate frames are warped to obtain coarse intermediate frames. Conditioned on the warped intermedia frames, a diffusion model is adopted to generate detailed and high-resolution target frames, which guarantees coherent and identity-preserved transitions. The cascaded conditional diffusion model decomposes the complex talking editing task into two flexible generation tasks, which provides a generalizable talking-face representation, seamless audio-visual transitions, and identity-preserved faces on a small dataset. Experiments show the effectiveness and superiority of the proposed method.
\end{abstract}

\section{Introduction}

\begin{figure}[]
    \centering
    \includegraphics[width=1.0\linewidth]{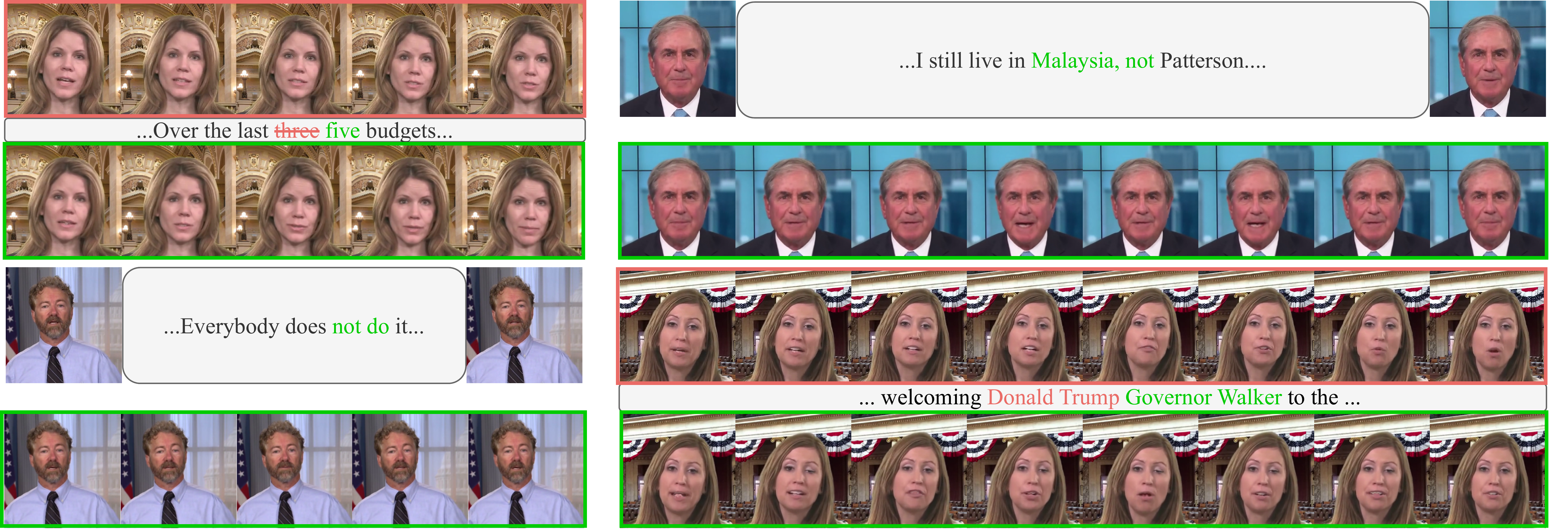}
    \caption{We implement talking-head video editing tasks in zero-shot scenarios, without the need for finetuning on specific character data. Frames marked in green are generated by our method. Our approach is not limited to word-level editing but also facilitates sentence-level editing.}
    \label{fig:example}
\end{figure}

Text-based talking-head video editing aims to efficiently insert, delete, and substitute segments of talking videos through a user-friendly text editing approach (Fig. \ref{fig:example}). It has a wide range of applications in video post-processing such as film-making, video advertising, and digital avatars. Unlike talking-head video generation \cite{ye2023geneface++,tian2024emo,xu2024vasa,wei2024aniportrait} that has been sufficiently explored, talking-head video editing remains under-explored. Previous text-based talking video editing methods~\cite{fried2019text,yao2021iterative} required extensive training on a large dataset of source videos to effectively edit videos while preserving the current character's identity. However, in real-world applications, users often need to edit only a short video clip without additional video data available.

Text-based talking-head video editing has several key challenges. \textit{\textbf{First}}, talking video editing is a complex learning task that requires lots of videos to learn a generalizable representation for video synthesis of any identity. Conventional one-shot methods like NeRF require extensive fine-tuning on specific individual video data. However, they still face issues with unnatural head poses and distortions between the head and torso.
\textit{\textbf{Second}}, it is difficult to achieve seamless transitions in lip movements, head gestures, and facial expressions between edited and unedited segments. \textit{\textbf{Third}}, during video editing, it is hard to maintain the consistency of a person's identity across adjacent frames.
Moreover, our method focuses on sentence-level editing instead of word-level, which is more difficult. We consider all of these challenges and propose a novel framework for talking video editing, as discussed in the following. 


Existing methods could offer a partial solution for talking video editing, which can roughly be categorized into two groups.
\textit{\textbf{In the first group}}, some text-based talking-head video editing  methods~\cite{fried2019text,yao2021iterative} search  similar phonemes from observed audios and blended the target talking videos by the corresponding video segments. However, it is difficult to extend to sentence-level talking editing and it requires at least minutes of source videos and hours of training time. Recently, a state-of-the-art text-based talking editing method \cite{yang2023context} proposed to utilize contextual information to train a non-autoregressive transformer on a large-scale video dataset and fine-tune the NeRF model on specific individual videos. The talking-face representation is not generalizable and it takes too much time for per-identity optimization on a long target video.
\textit{\textbf{In the second group}}, some talking generation methods~\cite{guo2021ad,liu2022semantic,shen2022learning} also partially inspire talking-head editing. They typically use NeRF~\cite{mildenhall2021nerf} to synthesize talking-head images with preserved identities after training on a small dataset. However, the head poses generated by these methods can only be sampled from the original pose sequences. For example, the Geneface series models~\cite{ye2023geneface,ye2023geneface++} render the head and torso separately using NeRF and then stitch them together. Although this can generate a variety of head poses, it results in unnatural artifacts at the head-torso junction. 
Some methods developed GAN-based models like Wav2Lip~\cite{wav2lip} and diffusion models, including AniPortrait~\cite{wei2024aniportrait}, EMO~\cite{tian2024emo}, and VASA~\cite{xu2024vasa}. However, they generally require extensive training on large datasets of audio-video data from various identities to successfully transfer to unseen identities. Moreover, it is unclear how to extend these models to talking-face editing which requires careful design to address coherent transitions between edited segments and unedited segments.

To address these issues, we propose a novel cascaded conditional diffused-based framework, which consists of two stages: audio to dense-landmark motion and motion to video. \textit{\textbf{In the first stage}}, we first propose a dynamic weighted in-context diffusion module to synthesize dense-landmark motions given an edited audio. The diffusion module circumvents traditional facial landmarks and instead generates landmark images. More precisely, given the context video frames, we extract the dense-landmark images with \cite{}. We train a diffusion by dynamically weighting the start frame and the end frame as conditions. \textit{\textbf{In the second stage}}, we introduce a warping-guided conditional diffusion module. The module first interpolates between the start and end frames of the editing interval to generate smooth intermediate frames. Then, with the help of the audio-to-dense motion images, these intermediate frames are warped to obtain coarse intermediate frames by a warping network \cite{}. Conditioned on the warped intermediate frames, a diffusion model is adopted to generate detailed and high-resolution target frames, which guarantees coherent and identity-preserved transitions. The cascaded conditional diffusion model decomposes the complex talking editing task into two flexible generation tasks, which provides a generalizable talking-face representation, seamless audio-visual transitions, and identity-preserved faces on a small dataset. 

Overall, our main contributions are summarized as follows: \textbf{(1)} We introduce a novel cascaded conditional diffusion framework for talking video editing. The framework decomposes the complex talking editing task into audio to dense-landmark motion and motion to video, which provides generalizable talking-face representation, significantly improves lip synchronization and head pose coherence, and maintains identity-preserved faces. \textbf{(2)} We propose a dynamic weighted in-context diffusion module to synthesize dense-landmark motions given an edited audio. The diffison model pioneers an audio to dense-landmark motion module that eschews traditional facial landmark keypoints in favor of landmark images. The method reduces the dependency on extra audio-video training data and ensures consistent identity preservation across the edited videos. \textbf{(3)} We propose a warping-guided conditional diffusion model for talking-head videos which is conditioned on image warping driven by dense-landmark motions. Guided by the warped talking-head images, the diffusion model yields detailed, coherent, and identity-preserved talking video frames. \textbf{(4)} Extensive experiments show the effectiveness and superiority of the proposed method.

\section{Related Work}
\textbf{Talking-head Video Generation.} Talking-head video generation aims to create talking videos of specific individuals based on their voices. The correlation between audio and facial features can be divided into verbal (related to lip movements) and non-verbal (involving head poses and expressions) components. Efforts focused on verbal generation aim to map audio signals to lip movements while keeping other areas of the face static. Such methods~\cite{wav2lip,shen2023difftalk,iplap} typically require a segment of source video and are limited by fixed head poses and expressions, which restricts the ability to generate rich facial details. Recent work~\cite{ye2024real3d,ma2023dreamtalk,he2023gaia} strives to create both verbal and non-verbal facial details. These methods typically operate in stages: first generating an intermediate representation of the face (such as landmarks~\cite{denselandmark}, 3DMM~\cite{ren2021pirenderer}, or mesh~\cite{wei2024aniportrait}) based on the audio signal, and then converting these intermediate representations into RGB videos using various rendering models (e.g., NeRF, Diffusion, GAN~\cite{facevid2vid}). Typically, the Geneface series models first generate 3D landmarks from the audio, which are then rendered into video using the NeRF model. Sadtalker~\cite{zhang2023sadtalker} generates a 3DMM intermediate representation through audio2headpose and audio2expression modules, which is then rendered into video by an Image Generator. More recently, models like EMO, AniPortrait, and VASA have leveraged the powerful generative capabilities of diffusion models to produce high-quality talking videos.  They utilize various motion representations and integration methods to extend the image generation capabilities of diffusion models to the video domain.

\textbf{Text-driven Talking Video Editing.} 
By performing local edits on the text (insert, delete, and substitute), the modified audio is obtained, which then drives the generation of the edited talking video. It is essential to ensure that unedited regions remain unchanged, that visual frames in the edited area are aligned with the audio, and that there is a smooth transition between adjacent frames. 
\cite{fried2019text} proposed a text-driven editing method based on a viseme-phoneme dictionary, their approach requires an hour of target person data and extensive training time. 
\cite{yao2021iterative} adopted an iterative talking-head video editing process and alleviated the issue of scarce target person data using neural redirection techniques. However, their method still requires 2-3 minutes of target person data for fine-tuning during the inference stage. Both of these methods are limited to editing the verbal component of the face.
\cite{yang2023context} made full use of the context information from the source video to implement talking video editing. Their framework is primarily divided into two parts: motion prediction and motion-conditioned rendering. They first extract motion from the context information within the edited interval, then input it into a NeRF-based renderer to synthesize the video frame sequence. Although this method addresses the problem of word-level talking video editing to some extent, it struggles with sentence-level editing and large-pose editing. In contrast, our approach considers both word-level and sentence-level editing. Additionally, because it relies on a NeRF-based renderer, it still requires fine-tuning on the renderer, which is time-consuming—about 25 minutes for a 15-second video.  In contrast, our method does not require fine-tuning on specific characters and is designed for zero-shot scenarios, significantly reducing the time required by half.

\textbf{Audio to Motion.}
Facial representations vary widely, including landmarks, 3DMM (blendershape), implicit representations, and so on. Landmarks explicitly represent facial features through keypoints, typically marking specific locations of the eyes, eyebrows, nose, mouth, and chin. Common configurations include the 68-point and 478-point landmarks~\cite{lugaresi2019mediapipe}. Predicting a landmark sequence from audio provides an intuitive representation of talking motion. These methods~\cite{ye2023geneface,ye2023geneface++} often require extensive training on large lip-reading datasets to accurately map audio signals to facial movements. Although they can achieve alignment between the generated landmark sequences and the audio, they fail to ensure the preservation of the individual's identity. To maintain the identity, it's common to incorporate reference information specific to the individual. MODA~\cite{liu2023moda} utilizes subject information, along with lip and eye priors as input. IP-LAP~\cite{iplap} uses the upper half of the face, excluding the lips, as input to maintain the identity. AniPortrait directly generates a mesh from a reference image and audio signals and then projects this mesh onto 3D landmarks. 
BlenderShape (3DMM) is another common form of 3D facial representation. Typically, it is necessary to run two parallel branches~\cite{ma2023dreamtalk,peng2023synctalk,sun2023vividtalk,zhang2023sadtalker} to predict head pose and facial expressions separately. These predictions are then coupled to provide a comprehensive 3D facial animation that incorporates both the expressions and the orientation of the head.
Besides explicit representations, there are also implicit ones. GAIA~\cite{he2023gaia} generates the motion latent sequence conditioned on the speech at each timestamp. DAE-Talker~\cite{du2023dae} trains a speech-to-latent model based on the Conformer architecture to predict the corresponding motion latent representations from speech. However, implicit representation methods often require joint training with the rendering model, which limits their ability to be fully decoupled and transferred. 


\section{Method}
\begin{figure}[!t]
    \centering
    \includegraphics[width=0.8\linewidth]{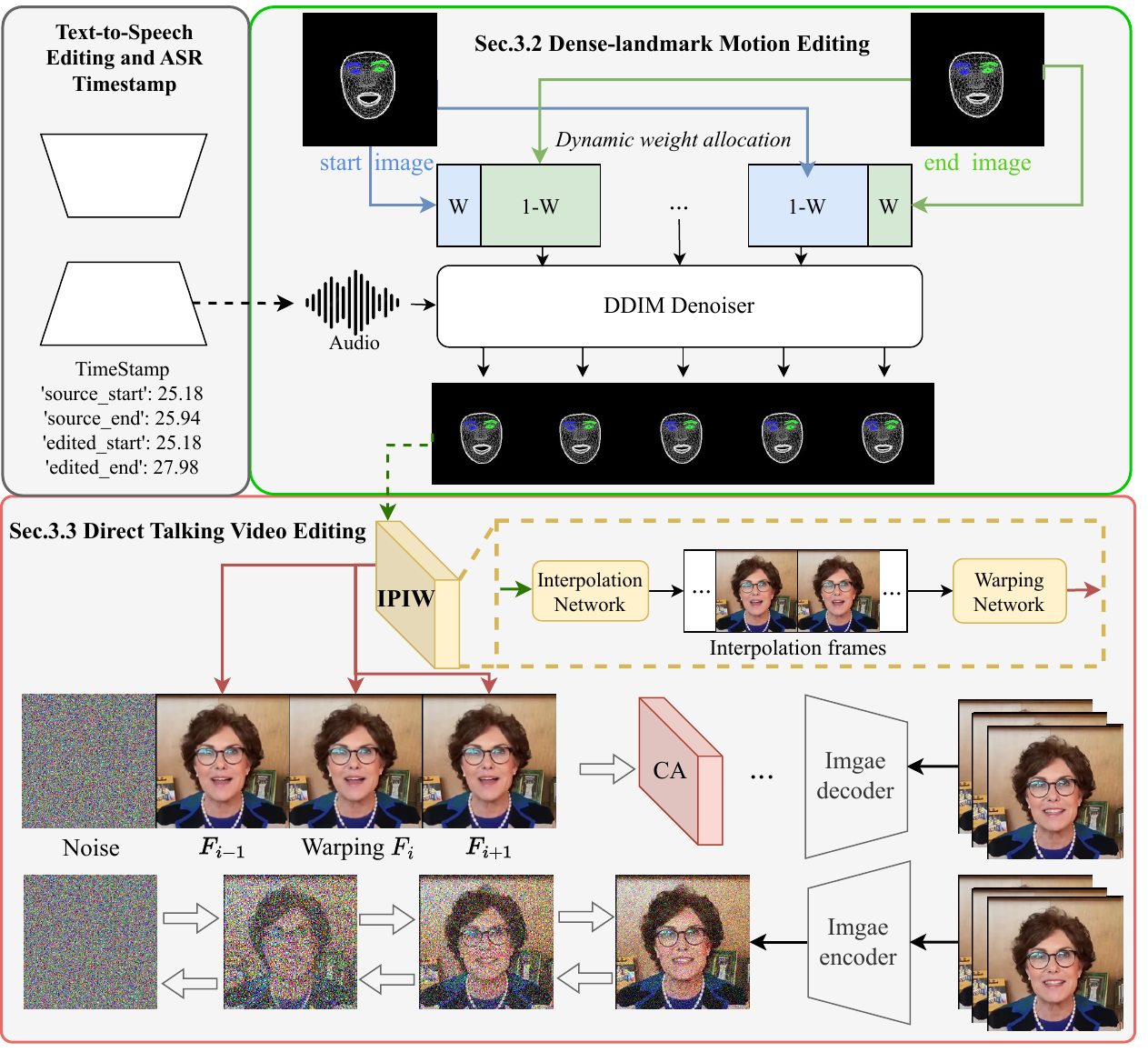}
    \caption{Overview of the proposed method. CA represents the cross-attention mechanism.}
    \label{fig:overall}
\end{figure}
The goal of talking video editing is to manipulate the talking-head video of an arbitrary person such that the edited video synchronizes an edited audio. Unedited segments should remain unchanged and transition smoothly with the edited regions. Specifically, given an original video \(\mathcal{V}\) with the corresponding audio \(\mathcal{A}\), we first edit the audio \(\mathcal{A}\) as \(\mathcal{A}^{'}\). Then we edit video \(\mathcal{V}\) as a video \(\mathcal{V}^{'}\) to synchronize \(\mathcal{A}^{'}\). Intuitively, several challenges exist in talking video editing. 
\textit{\textbf{First}}, talking video editing is a high-dimensional learning task that requires lots of videos to learn a generalizable representation for video synthesis of any identity. Conventional one-shot methods like NeRF require extensive fine-tuning on specific individual video data. In addition, they still face issues with unnatural head poses and distortions between the head and torso.
\textit{\textbf{Second}}, it is difficult to achieve seamless transitions in lip movements, head gestures, and facial expressions between edited and unedited segments. \textit{\textbf{Third}}, during video editing, it is hard to maintain the consistency of a person's identity across adjacent frames.
Moreover, our method focuses on sentence-level editing instead of word-level, which is more difficult. We consider all of these challenges and propose a novel framework for talking video editing, as discussed in the following. 

\subsection{Overview of the Proposed Framework}
The overview of the proposed framework is illustrated in Fig. \ref{fig:overall}. Basically, the framework has three main components, i.e., text-to-speech editing, dense-landmark motion editing, and direct talking video editing. 
Our editing operations include ``insert", ``delete", and ``substitute". Given an original video $\mathcal{V}$ with the corresponding audio $\mathcal{A}$, we first edit the audio $\mathcal{A}$ as $\mathcal{A}^{'}$ with a text-to-speech editing module~\cite{radford2023robust}. To alleviate the problem of limited video data for high-dimensional video editing, we propose to decompose talking video editing into low-dimensional dense-landmark motion editing and warping-guided video editing. The dense-landmark motion editing enables the framework to describe the principal component of the talking motion while the warping-guided video editing allows the framework to synthesize detailed identity-preserved talking faces conditioned on the dense-landmark motions.

\subsection{Dense-Landmark Motion Editing with Dynamic Weighted In-Context Diffusion Model}

Due to the weak correlation between speech and talking videos, conventional talking video generation methods require lots of talking videos to learn a good semantic-aligned representation for two modalities. To solve this problem, an intuitive way is to build a bridge between speech and talking videos, such as landmarks and 3D morphable models (3DMM)~\cite{3dmm}. Compared with 3DMM, the landmark representation can encapsulate elements of a person's identity, facial expressions, and head poses. Instead of using sparse landmarks \cite{sparseldm} (e.g., 68 keypoints), we use 468 keypoints for representation as discussed in Mediapipe~\cite{lugaresi2019mediapipe}, which provides richer facial details.

\paragraph{Dense-Landmark Representation.} We first propose a dense-landmark image representation that builds a bridge between audio and the landmark space. Unlike traditional methods where audio is used to generate landmark keypoint coordinates, our method predicts the landmark image directly, building a stronger relationship between the audio and the motion of a specific person. The proposed representation ensures a more reliable and identity-constrained generation of facial movements, resulting in smoother and more natural video frames. Using the generative paradigm of diffusion models, we generate landmark images within the editing interval. 
The dense landmark image inherently contains information about a person's identity. By incorporating the dense landmark images of the initial and final frames into the editing interval, our approach preserves the person's identity even in zero-shot scenarios and achieves smooth motion prediction and audio alignment. It thus addresses the challenge of creating identity-bound landmark images without the need for extensive paired data while maintaining continuity and realism in video editing tasks.

\paragraph{Conventional Image Diffusion Model.}
The latent diffusion model (LDM)~\cite{ldm} offers an efficient method for image generation. Based on a pair of pre-trained components—an image encoder \( \mathcal{E} \) and an image decoder \( \mathcal{D} \)—the input image \( I \in \mathbb{R}^{H\times W\times3}\) is encoded into a feature \( z_0 \in \mathbb{R}^{h\times w\times3} \) in the latent space, where \( H \) and \( W \) denote the height and width of the original image, respectively. \( h \) and \( w \) represent the dimensions downsampled to the latent space. LDM is thus constructed as a denoising network \(\theta\) operating in the latent space, designed to learn the reverse process of removing noise from the original data. The corresponding training objective can be expressed as follows:
\begin{equation}
    L_{obj} = \mathbb{E}_{z, \epsilon \sim \mathcal{N}(0,1), t} \left[ \left\| \epsilon - \theta(z_t, t) \right\|^2_2 \right]
\end{equation}
where \( t \in [1, 2, \ldots, T] \), and \( z_t \) is obtained by adding noise to \( z_0 \) over \( t \) steps through the forward process. The final denoising result is then mapped back to the original RGB image space through the pre-trained image decoder \( \mathcal{D} \).

\paragraph{Dynamic Weighted In-Context Diffusion Model.} Different editing methods produce varying audio outputs, with subtle variations in different speech patterns that can affect the smooth translation of video frames. To ensure a smooth transition between edited and unedited sections, we propose a context-aware conditional input to guide the generation of smooth and audio-aligned dense motion frames. This approach helps to create video frames that seamlessly transition between edited and unedited regions, reducing abrupt changes and maintaining visual continuity. Given that video frames exhibit a temporal correlation that diminishes with distance, we apply dual-linear interpolation to all frames within the editing interval. This involves linearly interpolating the frames based on their temporal distance from the interval's start and end frames. The dynamic weight in-context diffusion model can be formulated as
\begin{equation}
    L_{obj} = \mathbb{E}_{z, \epsilon \sim \mathcal{N}(0,1), t} \left[ \left\| \epsilon - \theta(z_t, t, I_{start}, I_{end}, C_{\mathcal{A}^{'}}) \right\|^2_2 \right].
\end{equation}
More precisely, the first frame within the interval has interpolation weights of \((bs-1)/bs\) and \(1/bs\), relative to the start and end frames, respectively. The approach ensures a smooth progression of frames and maintains temporal consistency within the editing interval.

\subsection{Direct Talking Video Editing with Warping-Guided Conditional Diffusion Model}

Directly driving a talking face with the generated landmarks ~\cite{ye2023geneface,ye2023geneface++} fails to capture the unique identity of specific individuals, leading to distorted and unnatural video frames in subsequent rendering. After we obtain edited dense-landmark motions, our goal is then to drive the talking video editing. To generate identity-preserved talking videos conditioned on unedited audio-video context, we first propose an ID-preserving interpolation-warping to generate coarse talking videos. Using the coarse talking videos as conditions, we then propose a warping-guided conditional diffusion model to generate high-quality identity-preserved talking videos.

\textbf{ID-Preserving Interpolation-Warping.}
The dense-landmark motion editing module produces dense landmark images aligned with the audio. However, using these images directly to render original video frames can cause distortions in the foreground and background. This issue is less pronounced with simple or solid-color background videos. To solve it, we employ a progressive rendering mechanism consisting of the \textit{IPIW} module and the warping-guided conditional diffusion module.

Within the editing interval, we only have information on the start and end frames. We first interpolate between the start and end frames to create smooth, ID-preserving intermediate frames. Then, using the dense landmark images from the interpolation frames and the target landmark images (obtained through the dense-landmark motion editing module), we utilize the optical flow to warp the interpolation frames. This process results in coarse target frames that are aligned with the audio. Since the entire procedure is centered around a specific individual, we achieve ID-preserved target frames. These frames provide robust supervisory information for the subsequent warping-guided conditional diffusion module. Specifically, we begin by utilizing a video frame interpolation method~\cite{huang2022rife} to generate intermediate frames. Subsequently, a warping network~\cite{zhang2023metaportrait} estimates the nonlinear transformation from the interpolation image to the target image. The warping network primarily consists of a motion encoder, an optical flow estimation network, and an image decoder. The motion encoder takes the interpolation landmark image \(I_s^{ldm}\) and the target landmark image \(I_t^{ldm}\) as inputs to produce the motion code \(z_m\). This motion code is then fed into the optical flow estimation network via an AdaIN~\cite{huang2017adain}, generating the motion flow relative to the interpolation frame. Finally, this motion flow is fed into the image decoder, which uses it along with the source frame \(I_t^{ldm}\) to perform the final warping and obtain the coarse target frames.

\textbf{Warping-Guided Conditional Diffusion Model.}  While the \textit{IPIW} module successfully maintains the identity in the warped target frames, it inevitably introduces artifacts and causes visual blurriness. Therefore, a refinement module is necessary to improve the visual quality of the target frames while preserving temporal consistency and identity. Diffusion models have been extensively validated for their high-quality image generation capabilities. However, in video generation tasks, strong temporal constraints are required to ensure consistency, as demonstrated by approaches like animatediff~\cite{guo2023animatediff}. Therefore, we designed a warping-guided conditional diffusion module to achieve this goal. Given that we have already obtained coarse video frames in the \textit{IPIW} module, we use these frames as conditional features in the denoising stage to ensure temporal consistency among the generated target frames. Additionally, by leveraging latent space learning techniques, we can achieve higher-resolution synthesis without incurring additional computational cost.

As depicted in Fig.~\ref{fig:overall}, we introduce the denoising network, which is based on the UNet architecture~\cite{unet} and supplemented with a cross-attention mechanism to improve multi-modal learning. The coarse target frames obtained from the \textit{IPIW} module are encoded into the latent feature \(Z_c\) using a pre-trained Vector Quantized network. To ensure the temporal consistency of the video frames generated by the image diffusion model, we concatenate the features of the preceding and following frames with the current frame's feature \(C_n\) to obtain \(C_{\mathcal{V}}\). Specifically, \(C_p^{*} = (C_p + C_n) * 0.5\) and \(C_f^{*} = (C_f + C_n) * 0.5\), where \(C_p\) and \(C_f\) represent the latent features of the before and after frames, respectively. \(C_p^{*}\) and \(C_f^{*}\) represent the updated processing frame features and following frame features, respectively, after fusion with the current frame's visual characteristics. \(C_{\mathcal{V}}\) is directly fed into the denoising network as the primary subject for denoising. Consequently, all conditioning information is incorporated into the denoising network to guide the refinement of the target frame, which is given by
\begin{equation}
    L_{obj} = \mathbb{E}_{z, \epsilon \sim \mathcal{N}(0,1), t} \left[ \left\| \epsilon - \theta(z_t, t, C_{\mathcal{A}^{'}}, C_{\mathcal{V}}) \right\|^2_2 \right].
\end{equation}
Note that the audio feature \(C_{\mathcal{A}^{'}}\) acts as the key and value in the cross-attention mechanism, engaging in multi-modal attention learning with \(C_{\mathcal{V}}\).

\section{Experiments}
\subsection{Experimental Settings}
\paragraph{Datasets.}
To train the talking-head video editing network, we used the audio-visual dataset HDTF~\cite{hdtf}. Since some original data is missing, we collected talking videos containing 172 different identities, totaling 580,486 image frames, equivalent to 3.62 hours. We implemented an identity-first data partitioning strategy to ensure that the identities in the test set were not previously seen by the model, thus evaluating the effectiveness of our method on unseen identities. To further demonstrate the superiority of our method, we used three different data-splitting strategies: (1) The training and validation sets together account for 50\%, and the test set accounts for 50\%. (2) The training and validation sets together account for 25\%, and the test set accounts for 75\%. (3) The training and validation sets together account for 12.5\%, and the test set accounts for 87.5\%. The training set and validation set are split in a 9:1 ratio.

\paragraph{Metric.} We primarily evaluate the performance based on five aspects: quality, consistency, smoothness, and subjective assessments.
(1) \textbf{Quality}: PSNR, SSIM~\cite{ssim}, and LPIPS~\cite{lpips} are three primary metrics for assessing the quality of generated video frames. (2) \textbf{Consistency}: The SyncNet score (distance \(\downarrow\) and confidence \(\uparrow\))~\cite{synet} measures audio-visual consistency, crucial for audio-driven talking video generation tasks.  (3) \textbf{Smoothness}: We introduce the metric F-SSIM to evaluate the smoothness of videos. This metric measures the SSIM between frames within transitional regions and editing intervals. (4) \textbf{User Study}: Additionally, we conduct a user study using the 5-Point Likert scale proposed by previous works~\cite{yang2023context}. We selected 28 different character IDs for evaluation, with each character undergoing three types of editing operations: delete, substitute, and insert, resulting in a total of 84 generated samples. The "mean" refers to the average score, while "real" represents the percentage of scores that are 4 and 5.

\paragraph{Network and Training Details.}
The dynamic weighted in-context diffusion model is designed based on the Latent Diffusion architecture, utilizing a DDIM denoiser~\cite{ddim} for accelerated denoising. The audio signals employ DeepSpeech features~\cite{deepspeech}, integrating the features of four preceding and following frames with the current frame's audio features to ensure smooth transitions. The Interpolation module uses pre-trained weights from the RIFEM model~\cite{huang2022rife}. Similarly, the warping-guided conditional diffusion model is also designed based on the Latent Diffusion architecture and uses DDIM for denoising, with the denoising step count \(T\) set to 200. The input images are resized to 256\(\times\)256 pixels, and a downsampling factor \(f\) of 4 is set.

\paragraph{Baseline Methods}
Due to the lack of open-source availability of text-based talking video editing methods most relevant to our work, we chose to compare our approach with the current state-of-the-art methods for talking-head video generation. These generation methods rely on audio to fully synthesize the video, lacking the capability to edit specific areas within the audio. Therefore, we assume that the audio in our test dataset has been pre-edited, and each method generates a talking-head video based on this audio segment. This ensures fairness in the experimental evaluation.

We selected DiffTalk, Geneface++, Sadtalker, and Aniportrait as baseline models to verify the effectiveness of our method. DiffTalk requires a video segment as input to stabilize the head movements and synthesizes the edited lip movements based on the audio. We trained the DiffTalk model using the same data setup and default model parameter configuration as our method. Geneface++ and Sadtalker both use NeRF-based rendering models to convert pose information derived from audio into images. For Geneface++, fine-tuning is performed on real-person videos from the HDTF dataset, with specific details provided in the supplementary materials. Aniportrait employs a diffusion architecture to simultaneously generate head poses and facial expressions. For Sadtalker and Aniportrait, we utilize the pre-trained models from their respective source repositories. 

\subsection{Comparison with Baseline Methods}
We conducted training under different data volumes, including 50\% of the total data, 25\% of the total data, and 12.5\% of the total data. These three data setups exhibit a containment relationship, where the train and validation sets in the 50\% data setup include those in the 25\% data setup. Conversely, the test set in the 50\% data setup is contained within the test set of the 25\% data setup. Our evaluation results on the HDTF dataset, compared with some representative baseline talking head synthesis methods, are shown in Table~\ref{tab:word}. In the image quality metrics SSIM, PSNR, and LPIPS, our method far surpasses other baseline models. The DiffTalk method comes closest to ours, which is due to its requirement for input videos with fixed head poses and lip masks during the generation process, providing it with rich prior information. Geneface++, having been fine-tuned on target individuals, ranks second only to DiffTalk in terms of generation quality among the baseline models.

The Sadtalker and Aniportrait methods exhibit weaker performance in image quality metrics, indicating lower fidelity to the Ground Truth in their editing tasks. Compared to DiffTalk, they share the same prior information as ours, which only includes information from the initial and final frames. This suggests that they are capable of generating a higher diversity in their outputs during editing tasks, naturally placing them at a disadvantage in this metric compared to DiffTalk and Geneface++. 
\begin{table}[!t]
    \centering
    \caption{The quantitative evaluation of talking-head video editing}
    \adjustbox{width=\columnwidth}{
    \begin{tabular}{c|c|c|c|c|c|c c} \toprule
        \multirow{2}{*}{Method} & \multirow{2}{*}{SyncNet \(\downarrow\uparrow\)} & \multirow{2}{*}{PSNR \(\uparrow\)} & \multirow{2}{*}{SSIM \(\uparrow\)} & \multirow{2}{*}{LPIPS \(\downarrow\)} & \multirow{2}{*}{F-SSIM \(\uparrow\)} & \multicolumn{2}{c}{User study \(\uparrow\)} \\ 
        \cmidrule{7-8}
        & & & & & & Mean & Real \\
        \midrule
         Ground Truth &6.92/8.06 &N/A &1.0 &0.0 &0.9566 &N/A &N/A\\
         DiffTalk (50\%) &11.80/2.93 &23.76 &0.9214 &0.1057 &0.9613 &2.812 &27.68 \\
         DiffTalk (25\%) &12.75/1.79 &23.51 &0.9196 &0.1066 &0.9612 &2.719 &26.37\\
         Geneface++ &8.70/5.48 &15.66 &0.6803 &0.2081 &0.9578 &2.645 &24.89 \\ 
         SadTalker &8.65/6.81 &9.77 &0.40 &0.45 &\underline{0.9739} &2.584 &22.52 \\ 
         AniPortrait &10.38/3.79 &13.38 &0.60 &0.30 &\textbf{0.9766} &2.612 & 21.73\\ \midrule
         Ours (50\%) &\underline{8.6246/5.9723} &\textbf{26.47} &\textbf{0.9375} &\textbf{0.0945} &0.9564 &2.937 &30.56 \\ 
         Ours (25\%) &8.6288/5.9729 &\underline{26.41} &\underline{0.9364} &\underline{0.0948} &0.9562 &2.945 &29.47\\ 
         Ours (12.5\%) &\textbf{8.5656/6.0137} &26.21 &0.9345 &0.0953 &0.9558 &2.859 &28.99\\  \bottomrule       
    \end{tabular}}
    \label{tab:word}
\end{table}

However, in the critical audio-visual consistency metric, the Synet score, Sadtalker and Aniportrait have made significant strides, with Sadtalker ranking first among all baseline methods. This demonstrates that although they are less suited for talking-head video editing tasks, they still maintain superior capabilities in generating video consistency. However, they are still slightly inferior to our method, further validating that even though we use the same prior information as the state-of-the-art models (Sadtalker, Aniportrait) — the initial and final frames — our method retains superiority in consistency and fidelity of video editing.

\subsection{Ablation Study}
\begin{figure}[!t]
    \centering
    \includegraphics[width=0.9\linewidth]{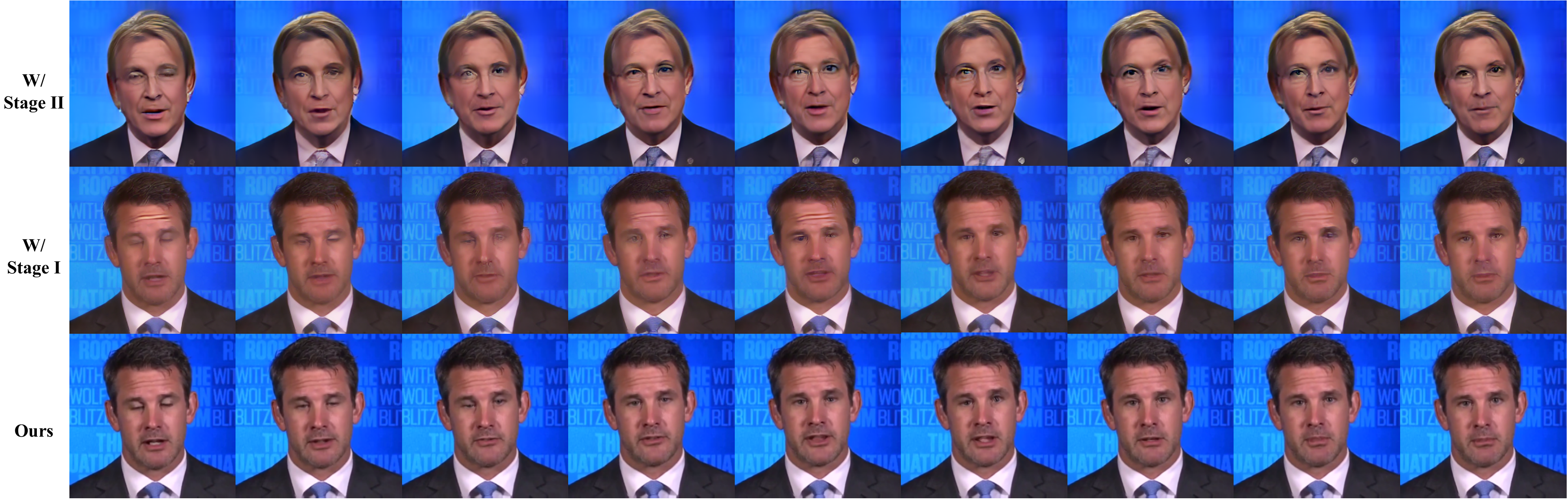}
    \caption{Visualized results of ablation study.}
    \label{fig:ablation}
\end{figure}

We conduct the ablation study across different rendering stages. Specifically, Stage I represents the ID-Preserving Interpolation-Warping module, while Stage II denotes the warping-guided conditional diffusion model.
W/StageI indicates that the output is the result generated by the Stage I ID-Preserving Interpolation-Warping module. W/StageII means that we directly encode the dense landmark image obtained from the dense-landmark motion editing model as conditional information into the warping-guided conditional diffusion model (no warp, only encode dense landmark image) to guide the rendering of RGB video frames. We conduct ablation experiments across the three aforementioned data configurations, and the results demonstrate the effectiveness of our progressive rendering network under any data setup.
As shown in Fig.~\ref{fig:ablation}, our method can achieve better image generation quality, audio-visual consistency, and smooth transitions. If rendering is conducted using only the Stage II process, where the dense landmark image obtained from the dense-landmark motion editing model is encoded into Stage II, the resulting video still struggles to maintain the person's identity, even when a reference image is added as an identity cue.
For more objective comparisons, we further evaluate the quantitative results in Table~\ref{tab:ablation}
\begin{table}[!t]
    \centering
    \caption{The quantitative evaluation of ablation study}
    \adjustbox{width=0.85\columnwidth}{
    \begin{tabular}{c|c|c|c|c|c} \toprule
         Method&  SyncNet \(\downarrow\uparrow\) & PSNR \(\uparrow\) &SSIM \(\uparrow\) &LPIPS\(\downarrow\) &F-SSIM\(\uparrow\) \\ \midrule
         W/ stage I (25\%) &9.1666/5.1830 &22.29 &0.8793 &0.1318 &0.9200 \\
         W/ stage I (12.5\%) &9.2257/5.1515 &23.01 &0.8904 &0.1289 &0.9468 \\ \midrule
         
         W/ stage II (25\%) &10.7537/5.7679 &6.06 &0.3633 &0.5754 &0.8634 \\ \midrule
         Ours (50\%) &\underline{8.6246/5.9723} &\textbf{26.47} &\textbf{0.9375} &\textbf{0.0945} &\textbf{0.9564} \\ 
         Ours (25\%) &8.6288/5.9729 &\underline{26.41} &\underline{0.9364} &\underline{0.0948} &\underline{0.9562} \\   
         Ours (12.5\%) &\textbf{8.5656/6.0137} &26.21 &0.9345 &0.0953 &0.9558 \\ \bottomrule
    \end{tabular}}
    \label{tab:ablation}
\end{table}

\section{Conclusion}
In conclusion, we present a novel cascaded conditional diffusion-based framework for text-based talking-head video editing, addressing three fundamental challenges: generalizable representations, seamless transitions, and identity preservation. Through the innovative use of a dynamic weighted in-context diffusion module and a warping-guided conditional diffusion module, our method significantly reduces the need for extensive video training data, simplifies the editing process, and enhances the quality and consistency of the edited videos. The experimental results demonstrate the effectiveness of our approach, showcasing its capability to perform robust video editing.
\paragraph{Limitation} Our method has some limitations. 
(1) Editing diversity: Text-based editing is not limited to the linguistic content of the actors. Modifying video style, character hair, lighting, and other visual elements based on text is also a promising direction for future exploration in text-based talking-video editing.
(2) Audio editing quality: Although the current TTS method produces highly realistic audio, an adequate solution for localized editing has yet to be found. This results in a domain shift in audio features, which in turn negatively impacts the quality of video editing. 
\paragraph{Broader Impacts} The edited talking-head video of Direct-Your-Talking might be misused to create misleading content or fake media. It can also edit personal imagery, leading to privacy concerns.

{
\small
 \bibliographystyle{plainnat}
 \bibliography{main}
}

\appendix



\newpage

\end{document}